\pdfoutput=1
\documentclass[11pt]{article}

\usepackage[final]{acl}
\usepackage{times}
\usepackage{latexsym}

\usepackage[T1]{fontenc}
\usepackage[utf8]{inputenc}

\usepackage{microtype}

\usepackage{inconsolata}

\usepackage{graphicx}

\title{Data Augmentation Integrating Dialogue Flow and Style to Adapt \\ Spoken Dialogue Systems to Low-Resource User Groups}

\author{Zhiyang Qi \\
  The University of \\ Electro-Communications \\
  1-5-1, Chofugaoka, Chofu, \\
  Tokyo, Japan \\
  \texttt{qizhiyang@uec.ac.jp} \\\And
  Michimasa Inaba \\
  The University of \\ Electro-Communications \\
  1-5-1, Chofugaoka, Chofu, \\
  Tokyo, Japan \\
  \texttt{m-inaba@uec.ac.jp} \\}

\begin{document}
\maketitle
\begin{abstract}
This study addresses the interaction challenges encountered by spoken dialogue systems (SDSs) when engaging with users who exhibit distinct conversational behaviors, particularly minors, in scenarios where data are scarce. We propose a novel data augmentation framework to enhance SDS performance for user groups with limited resources. Our approach leverages a large language model (LLM) to extract speaker styles and a pre-trained language model (PLM) to simulate dialogue act history. This method generates enriched and personalized dialogue data, facilitating improved interactions with unique user demographics. Extensive experiments validate the efficacy of our methodology, highlighting its potential to foster the development of more adaptive and inclusive dialogue systems.
\end{abstract}

\section{Introduction}

\begin{figure}[t!]
  \centering
  \includegraphics[width=\linewidth]{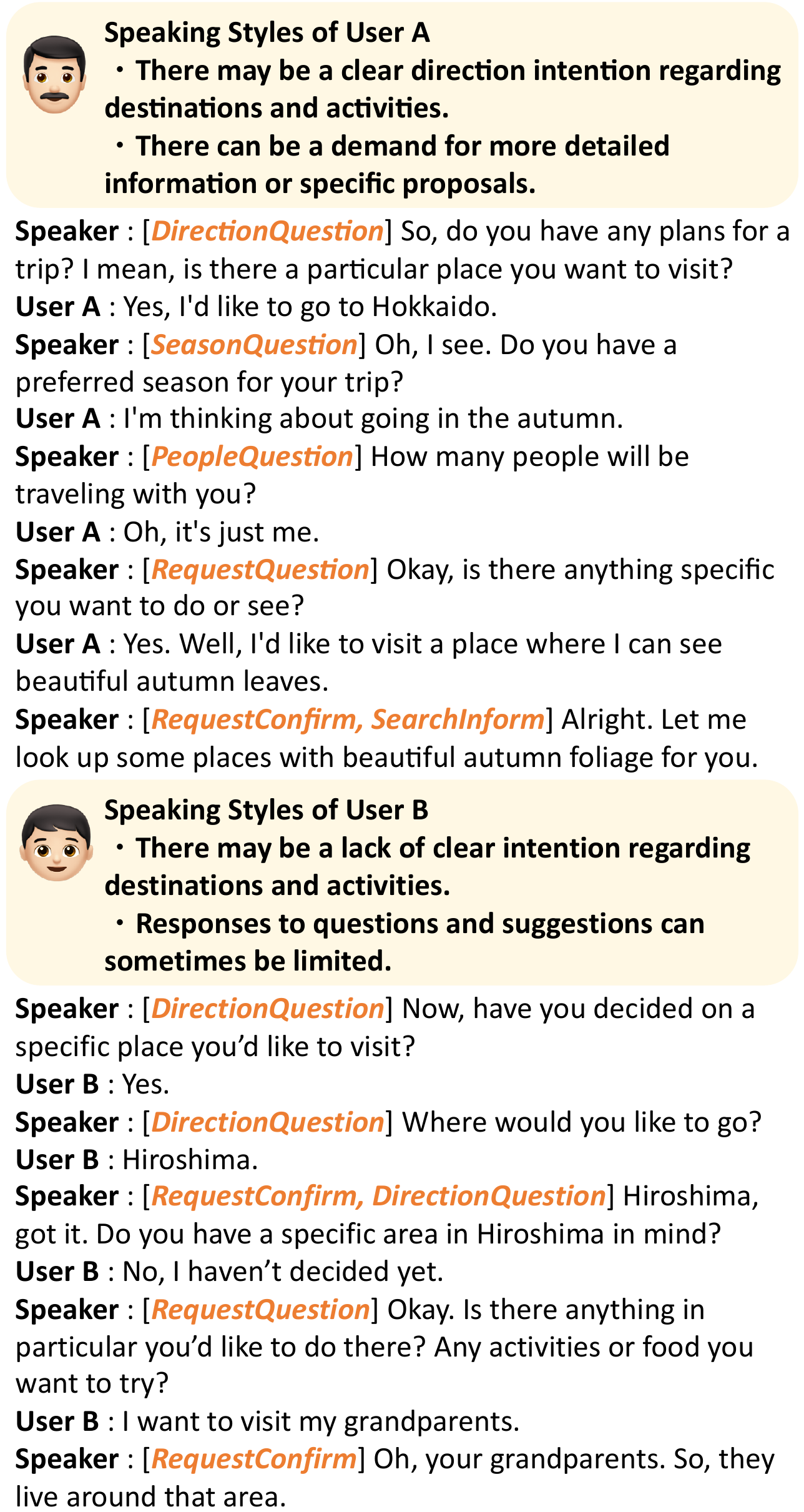}
  \caption{Real human-to-human conversations. Speakers adopt various dialogue acts when interacting with users employing diverse speaking styles.}
  \label{fig:pic_intro}
\end{figure}

As an innovative technology at the forefront of artificial intelligence and speech processing, spoken dialogue systems (SDSs) have attracted significant interest from both academia and industry \cite{Kawahara2018SpokenDS, si2023spokenwoz, Abdul-Kader2015, Kim2021HowRR}. 
Despite the powerful capabilities of large language models (LLMs), traditional SDS remain a focal point of research due to their superior control and interpretability \cite{2402.01761}. 
These systems are predominantly trained using data from human-to-human interactions, which highlight varying speaking styles, such as clarity of intentions, as depicted in Figure~\ref{fig:pic_intro}. This variability necessitates that human speakers adjust their dialogue strategies when engaging with different users. For instance, compared to adults, minors often exhibit less clarity in their intentions and give ambiguous responses, requiring more confirmatory language or additional inquiries to better adapt to the unique speaking styles of younger users. This adaptive approach is crucial for enhancing the effectiveness and user-friendliness of SDS in real-world scenarios.

However, adapting SDSs to these distinctive speaking styles typically requires a wealth of annotated dialogue data, which can be challenging to obtain in abundance owing to the minority status of users employing unique conversational strategies or behaviors. To enhance the performance of dialogue systems when engaging with data-scarce user groups exhibiting distinct speaking styles, targeted data augmentation is imperative, enabling the system to better cater to their needs.

This study introduces a tailored data augmentation framework designed specifically for low-resource user groups exhibiting distinctive conversational behaviors. Recognizing the unique conversational behaviors and challenges associated with minors and the inherent difficulty in obtaining their data \cite{difficult}, our study conducts experiments utilizing dialogue data from minors to facilitate targeted data augmentation for this demographic.

As depicted in Figure~\ref{fig:pic_intro}, the unique speaking style of users directly influences the speaker's dialogue acts (DAs) and indirectly shape response content. Therefore, our data augmentation framework focuses on the speaking styles of users and the trajectory of DAs. 

Specifically, we utilized a LLM to extract the speaking styles of such users and speakers interacting with them.
We then fine-tuned a pre-trained language model (PLM) using all available data in a low-resource setting to create varied histories of DAs for speakers interacting with these user groups. The resulting speaker styles and DA histories were input into the LLM to produce customized training dialogue data for these users. The primary goal is to enhance the model's ability to predict DAs when interacting with low-resource groups with unique speaking styles, as controlling the content of generated responses through DAs is deemed effective \cite{SeiyaKawano202136-4_E-KC9}.

This study's contributions are outlined below.

\begin{itemize}
\item We introduced a data augmentation method to enhance the performance of the DA prediction model when dealing with users who have limited data and unique conversational behaviors and styles.
\item Through multiple experiments conducted in a low-resource setting, we have discovered that the difficulty of DA prediction varies across different users and demonstrated the adaptability and effectiveness of our proposed method.
\end{itemize}

\begin{figure*}[t!]
  \centering
  \includegraphics[width=1\textwidth]{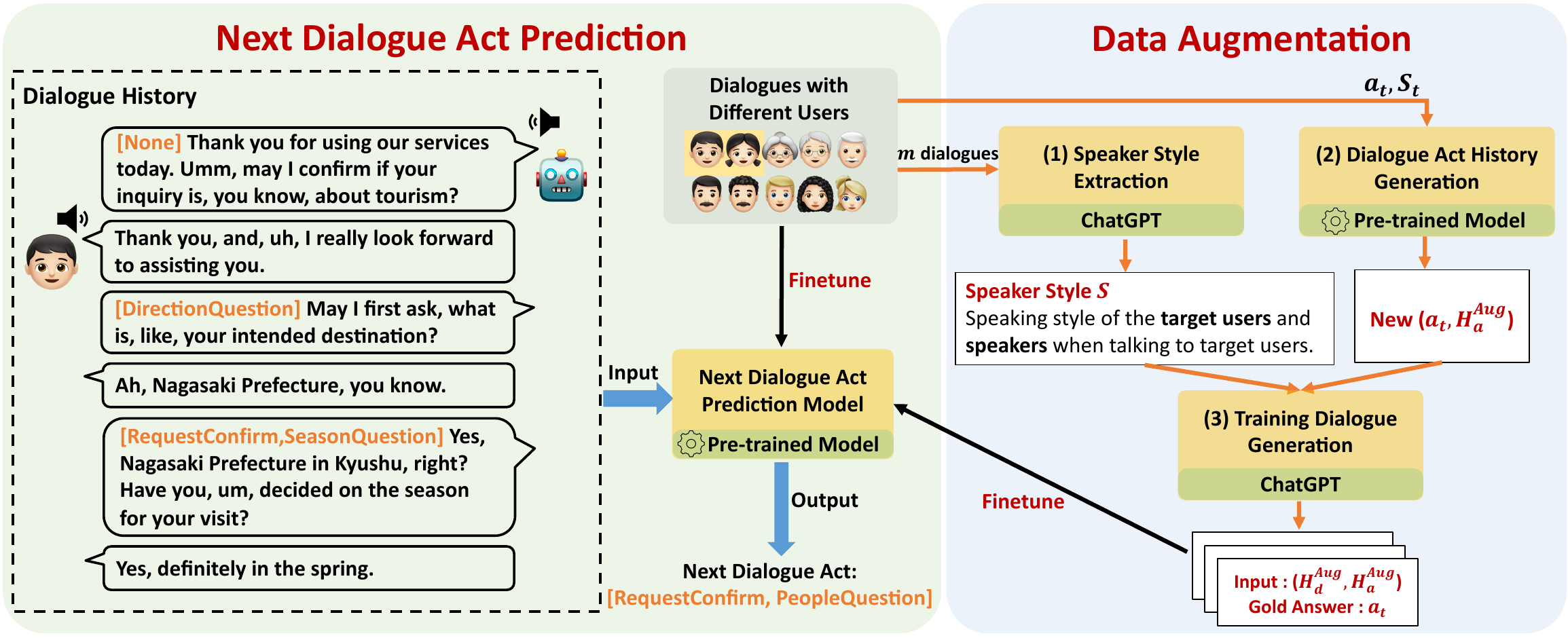}
  \caption{Our data augmentation framework is designed to improve the performance of the PLM in predicting DA when interacting with low-resource users who exhibit unique speaking styles. Beginning with dialogues that involve specific target users, we: (1) extract speaker styles, (2) generate DA histories of system interactions with these users, and (3) input this information into ChatGPT for tailored data augmentation.}
  \label{fig:our_model}
\end{figure*}

\section{Related Work}

The scarcity of annotated data and the challenge of data imbalance are persistent issues in various artificial intelligence domains \cite{Shorten2019ASO, rs12101688, AHMAD2021125834, hedderich-etal-2021-survey, kim-etal-2023-soda}. To address these effectively, data augmentation techniques have been employed, as demonstrated in prior research across different tasks \cite{feng-etal-2021-survey, 10.1145/3544558}. 
For instance, \citet{schick-schutze-2021-generating} generated text similarity datasets from scratch by instructing a large PLM. Similarly, \citet{liu-etal-2022-data} and \citet{chen-yang-2021-simple} enhanced data by manipulating individual utterances within dialogues—such as adding, deleting, changing their order, or regenerating them—while preserving the original meaning, which improved model performance in dialogue summarization tasks. While the abovementioned methods focus on generating individual sentences, our study aims to create coherent dialogues comprising multiple sentences tailored for specific target groups. 

% Biswesh Mohapatra等人使用预先训练的语言模型GPT2创建用户机器人和代理机器人，通过机器人交互的方式创建完整的任务指向型对话，并在MultiWOZ数据集和PersonaChat数据集的低资源设置上实现了显著改进。最近，由于LLM出色的文本生成能力，有些研究使用LLM进行数据增强。Yan Pan等人通过域外对话中抽取到的对话路径，利用LLM生成符合对话路径的域内任务指向型对话。这项研究的对话路径的概念和我们研究中的对话行为历史很像，但不同的是，他们是从现有的数据中抽取对话路径，而我们是根据现有数据生成新的对话行为历史，并且针对目标用户人群进行了优化。

\citet{mohapatra-etal-2021-simulated-chats} utilized GPT-2 \cite{Radford2019LanguageMA} to develop user and agent bots, generating comprehensive task-oriented dialogues through bot interactions, demonstrating notable enhancements in low-resource scenarios with datasets MultiWOZ \cite{budzianowski-etal-2018-multiwoz} and PersonaChat \cite{zhang-etal-2018-personalizing}. Recently, with the advanced text generation capabilities of LLMs, researchers have started using LLMs for data augmentation \cite{pan-etal-2023-data, kim-etal-2023-soda, wang-etal-2023-improving-conversational}. For instance, \citet{kim-etal-2023-soda} guided LLMs to generate a broad spectrum of social dialogues using social commonsense knowledge from a knowledge graph. \citet{pan-etal-2023-data} generated domain-specific, task-oriented dialogues by extracting dialogue paths from out-of-domain conversations. The concept of dialogue paths in their work aligns with the concept of DA history in our research. However, the key distinction is that while they extract DA paths from existing data, we generate tailored DA histories based on existing data, specifically optimized for target user groups.

% Shester Gueuwou et al. improved translation quality by clustering similar high-resource sign languages and incorporating them into the training of low-resource languages\cite{gueuwou-etal-2023-jwsign}.

% \section{Task and Dataset}

% \subsection{Task Definition}

% 预测行为标签任务如图3的左半部分所示，其定义如下。对话人为提供服务的operator和customer。假设当前回合为t，我们将前n个回合的对话历史（uo,uc）以及前n个回合的operator的对话行为历史（a）作为输入，输出为当前回合的对话行为at。

% The task of predicting dialogue act, as shown in the left half of Figure~\ref{fig:our_model}, is defined as follows. The participants in the dialogue are operator and customer. Assuming the current turn of the dialogue is turn $t$, we take the dialogue history from the previous $n$ turns $H_u=(u_{t-1}^o, u_{t-1}^c,...,u_{t-n}^o, u_{t-n}^c)$ along with the operator's dialogue act history $H_a=(a_{t-1}^o,...,a_{t-n}^o)$ from these turns, as the input. The output is the dialogue act $a_t^o$ for the current turn.

\section{The Proposed Framework}

In this study, we aim to enhance the DA prediction performance of the system when dealing with low-resource user groups that exhibit unique dialogue strategies, by generating training data through the proposed data augmentation framework. In the construction of SDSs, accurate DA prediction is crucial as it facilitates dialogue state tracking and guides response generation, thereby reducing erroneous responses \cite{10.1145/3166054.3166058}. 
The task depicted in the left portion of Figure~\ref{fig:our_model} is defined as follows.
Assuming the current turn of the dialogue is turn $t$, we utilize the dialogue history $H_d=(S_{t-n}, U_{t-n},...,S_{t-1}, U_{t-1})$ from the previous $n$ turns, along with the system's DA history $H_a=(a_{t-n},...,a_{t-1})$ from these turns, as the input. The output is the system's DA $a_t$ for the current turn.

Since we predict the current turn's DA based on the dialogue history and the system’s DA history, it becomes crucial to generate dialogue and system DA histories that closely align with the target user group. To achieve this, we control the generation of dialogue data by capturing the speaking style of dialogue participants and generating dialogue flows that mimic real human interactions with the target user group. The importance of this approach lies in the fact that the model can effectively understand and adapt to unique dialogue strategies only when the training data realistically simulates complex dialogue scenarios. In real human interactions, users with unique dialogue strategies are in the minority and exhibit considerable diversity. Due to the limitations in data scale, traditional training datasets often fail to cover this diversity, which limits the model's adaptability and accuracy when dealing with such users. By simulating the dialogue styles and processes of specific user groups, we can generate more diverse and precise training data, thereby enhancing the model's generalizability and adaptability to diverse users.

As illustrated in Figure~\ref{fig:our_model}, our data augmentation framework comprises three components: (1) employing ChatGPT\footnote{https://openai.com/blog/chatgpt} to extract the speaker's styles $S$, (2) finetuning a pre-trained model to generate the system's DA history $H_a^{Aug}=(a_{t-n}^{Aug},...,a_{t-1}^{Aug})$, and (3) inputting the extracted speaking styles $S$ and the generated system's DA history $H_a^{Aug}$ into ChatGPT to generate the training dialogue data $H_d^{Aug}=(S_{t-n}^{Aug}, U_{t-n}^{Aug},...,S_{t-1}^{Aug}, U_{t-1}^{Aug})$.

\subsection{Speaker Styles Extraction}
% \begin{figure}[t!]
%   \centering
%   \includegraphics[width=\linewidth]{pic_prompt_1.pdf}
%   \caption{Prompt for Speaker Style Extraction.}
%   \label{fig:pic_style}
% \end{figure}

Since the unique speaking styles employed by the target user group significantly influence the content of conversations, it's crucial to capture the speaking styles of this group by comparing dialogues from the target user group with those from non-target groups. This helps guide the subsequent generation of dialogues specifically tailored to the target user group. To facilitate this, we employ ChatGPT to extract speaker styles from conversations involving target users. 
% The prompt is illustrated in Figure~\ref{fig:pic_style}.

Specifically, we input a set of $m$ dialogues, half of which involve users from the target group and the other half from non-target user groups. This balanced approach allows for an effective comparison, helping to identify and differentiate prominent speaking characteristics unique to the target group. 
Subsequently, ChatGPT is utilized to generate outputs representing the speaking style of the target users, as well as the speaking style of speakers when engaging with the target user group. Notably, our primary focus is on extracting abstract styles, such as "target users often exhibit ambiguous intentions towards destinations and activities." These styles are crucial because they significantly influence the direction of the dialogue, thereby enhancing the realism and relevance of the generated dialogues to actual human conversations. 
The prompt and extracted speaker styles are presented in Appendix~\ref{sec:appendix_c}.

\subsection{DA History Generation}

% 如图所示，由于目标群体采用独特的对话策略，与他们互动的人的对话行为也会受到影响。我们现阶段的目标是生成多样化的、真实的对话行为历史记录，并专门针对具有独特说话策略的群体进行优化。为此，如图3所示，我们利用现有数据对 PLM 进行微调，生成前n轮的对话行为历史记录。

\begin{figure}[t!]
  \centering
  \includegraphics[width=1\linewidth]{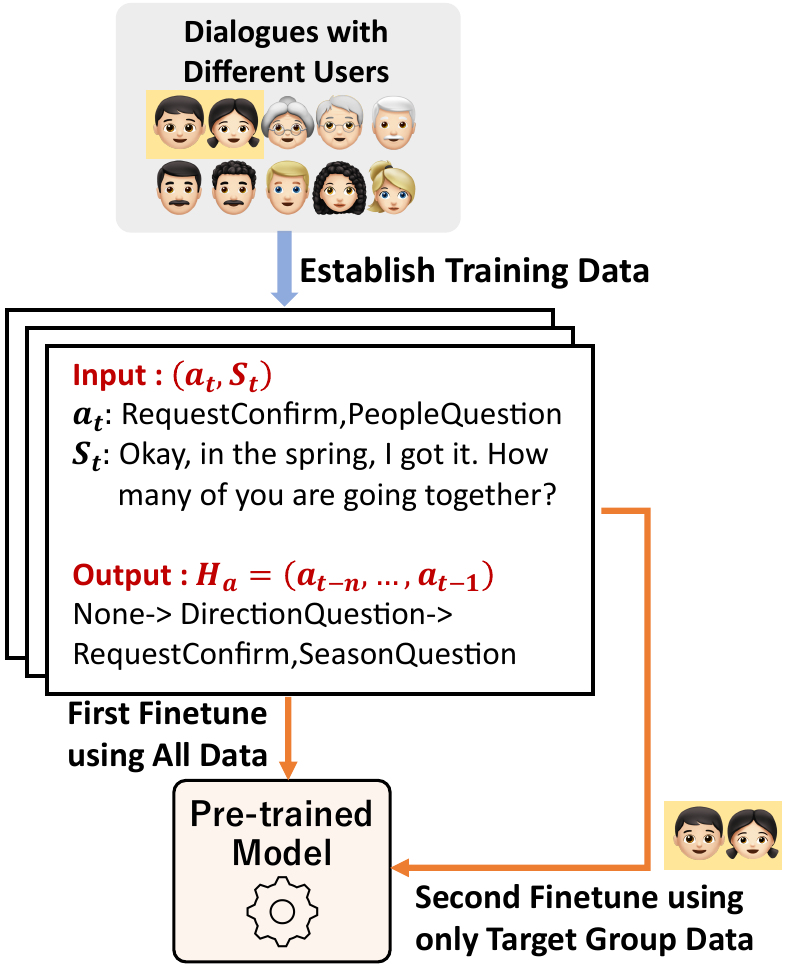}
  \caption{DA History Generation. We conduct two rounds of finetuning: the first round using all available data, and the second round using only data from the target user group, to ensure the generated DA history more closely aligns with the target demographic.}
  \label{fig:pic_da_gen}
\end{figure}

As depicted in Figure~\ref{fig:pic_intro}, the unique conversational strategies employed by the target group also significantly influence the DAs of those engaging with them. Our objective at this stage is to generate a diverse and realistic DA history $H_a^{Aug}$ that is specifically optimized for groups with distinctive speaking strategies. As shown in Figure~\ref{fig:pic_da_gen}, we achieve this by finetuning a PLM using existing data to generate the system's DA history $H_a^{Aug}$ for the previous $n$ turns.

% 具体来说，我们利用当前回合 $t$ 的 DA $a_t$ 和语篇 $S_t$ 作为输入，并利用前 $n$ 个回合的 DA 历史记录 $H_a$ 作为输出，建立训练数据。然后将这些数据分为两组：一组用于训练，另一组用于生成。我们首先使用所有可用的训练数据对 PLM 进行微调，以捕捉与真实人类对话非常相似的 DA 历史记录。随后，我们利用专门来自目标用户群的训练数据进行二次微调。这种双重微调方法确保了模型能够生成模仿真实人类对话的会话记录，并与目标用户的独特说话策略保持一致。因为在第一次微调中我们使用了大量的数据，因此模型能够生成模仿真实人类对话的DA历史，而第二次我们使用少量的目标用户群体的数据微调，可以使模型生成的DA历史更偏向目标用户群体。

In particular, we utilize the DA $a_t$ and utterance $S_t$  from the current turn $t$ as inputs, with the DA history $H_a$ from the previous $n$ turns as the desired output to establish training data. These data are then divided into two sets: one for training and the other for generation. Initially, we finetune the PLM using all available training data to capture DA histories that closely resemble real human conversations. Subsequently, we conduct a secondary finetuning utilizing training data exclusively from the target user group. This dual finetuning approach ensures that the model can generate DA histories that closely mimic real human dialogues and align with the unique speaking strategies of the target users. The first finetuning, which employs a relatively large dataset, enables the model to produce DA histories that mirror authentic human interactions. The second finetuning, focused on a smaller dataset specific to the target user group, allows the model to better tailor the DA histories to their unique characteristics.

% 在生成阶段，我们将当前回合的店员角色的说话行动和话语作为输入，输出前t个回合的店员角色的说话行动历史。为了保证多样性，我们一次生成多个输出，仅保留之前没有出现过的(a,a)组合。

During the generation phase, we input the the DA $a_t$ and utterance $S_t$ from the current turn $t$ and generate the DA history $H_a^{Aug}$ from the previous $n$ turns. To ensure diversity, we simultaneously generate multiple outputs, selecting only those $(a_t,H_a^{Aug})$ combinations that have not been previously observed.

\label{sec:dahg}

\subsection{Dialogue Generation}

% 现在我们已经生成了符合这类用户对话的说话人的特征和对话行为历史，在这一阶段，我们希望生成符合特征和对话行为历史的对话。
% 考虑到ChatGPT在对话生成方面的优越性能，我们在这一阶段使用ChatGPT生成训练用的对话数据。我们将提取到的说话人特征和生成的对话行动历史输入ChatGPT，利用图4的few-shot prompt引导其生成符合未成年风格的对话。最终，我们将生成的对话和对话行动历史作为输入，a0作为黄金回答来构建训练数据。其目的是提高模型在面对这类拥有特殊说话策略且数据稀缺的未成年人时的对话行为预测性能。

Having obtained speaker styles and DA history tailored to users employing unique dialogue strategies, our ultimate goal is to generate dialogues corresponding to these styles and histories to enrich the training data for DA prediction. 
At this stage, we leverage ChatGPT's powerful generation capabilities to create dialogue data for training purposes.
Utilizing a few-shot prompt, we input the extracted speaking styles $S$ and the DA histories $H_a^{Aug}$ into ChatGPT to generate dialogues $H_d^{Aug}$ that reflect the conversational style of the target users. Subsequently, we use the generated dialogues $H_d^{Aug}$ and DA histories $H_a^{Aug}$ as inputs, with $a_t$ as the gold-standard answer, to construct the training data. The prompts used for generating these dialogues are detailed in Appendix~\ref{sec:appendix_d}.

This approach aims to enhance the model's ability to predict DAs when interacting with target users who exhibit unique conversational strategies. It effectively addresses the challenge of data scarcity by employing data augmentation.

% as depicted in Figure~\ref{fig:pic_prompt},

% Having generated the speaker styles and DA history that align with such users who use unique conversational strategies, we now aim to generate dialogues that correspond with these styles and histories.
% Considering ChatGPT's superior capabilities in dialogue generation, we use it to generate training data at this stage. We input the extracted speaking styles $S$ and the generated DA histories $H_a$ into ChatGPT, using a few-shot prompt as shown in Figure~\ref{fig:pic_prompt} to guide it in generating dialogues $H_d$ that conform to the style of target users. Ultimately, we use the generated dialogues $H_d$ and DA histories $H_a$ as inputs, with $a_t$ as the gold answer, to construct the training data. Thereby enhancing the model's performance in predicting DA when interacting with target users who have unique speaking strategies and scarce data through data augmentation.

\section{Experiment}
% 为了评估所提出的数据扩增框架的有效性，我们使用未成年人的数据进行了实验，在数据集的实际对话中他们使用了独特的对话风格和策略。我们在低资源环境下进行了多次分拆实验，每次分拆都使用了不同的未成年人数据子集。我们在不同规模的数据集上训练了多个DA预测模型，包括使用添加到现有数据集的增强数据训练的模型。

To evaluate the effectiveness of the proposed data augmentation framework, we conducted experiments using data from minors who employed unique conversational styles and strategies in actual dialogues within the dataset. These experiments were carried out in a low-resource setting across multiple splits, each utilizing different subsets of data from minors. We trained multiple DA prediction models on datasets of varying sizes, including models trained with augmented data added to the existing datasets.

\subsection{Dataset}

% 在本研究中，我们使用了一个名为“旅行社任务”的多模态对话数据集，因为它收集了来自不同年龄段用户的对话，且对话行为被详细注释。这个数据集包含了115小时、共330个对话，每个对话持续约20分钟。对话内容关于对不同年龄段的用户进行旅游目的地推荐。
% 数据集作者通过使用隐马尔可夫模型（HMM）对注释的DA系列进行建模，以分析不同年龄段的对话过渡有何不同。一个显著的发现是，与其他年龄组的用户相比，未成年用户在对话中展现出了独特的策略。这些年轻的对话者往往不太倾向于表达独立观点，并展现出与成年人不同的对话模式。
% 这些对话是通过 Zoom 视频通话进行的，涉及 6 名操作员和 55 名客户。客户包括 20 名未成年人、25 名成年人和 10 名老年人。每位客户参与了 6 次对话。DAs 注释是按segment为单位进行的，segment是比话语还要小的每个segment被注释为包含None在内的29个事先定义好的对话行为中的一个。tag的详情在附录A展示。
% 由于被标记为None的segment通常是“Yeah”或“Uh-huh”等内容，而我们的目的是通过DA tag控制对话系统产生准确且有意义的回答，因此在本研究中，我们仅保留黄金答案不为None的训练实例。此外，我们在实验中使用基于文本形式的人工转录而非录音数据。

This study utilized a multimodal dialogue Japanese dataset known as the “Travel Agency Task Dialogue Corpus” \cite{inaba-etal-2022-collection, inaba-etal-2024-collection}, which features conversations from users of various age groups, with detailed annotations of DAs.
This dataset contains 115 hours of dialogue, spanning 330 conversations, with each averaging about 20 minutes. The dialogues were facilitated via Zoom video calls, involving six operators and 55 customers, including 20 minors (ages 7-17), 25 adults (ages 20-60), and 10 seniors (ages 65-72). Each customer participated in six dialogues.

The dialogues revolve around recommending travel destinations to users across various age groups. The dataset authors employed a Hidden Markov Model (HMM) \cite{18626} to analyze the transitions in dialogue among different age groups using sequences of DAs. 
A notable observation was that minors often used unique dialogue strategies compared to other age groups, typically expressing fewer independent opinions. The annotation of DAs was performed by functional segment, a unit smaller than an utterance. Each operator's segment is annotated as one of the 28 predefined DAs related to travel destination recommendations, or as "None". Examples of these DAs include asking about the travel season (SeasonQuestion) and summarizing the travel plan (TravelSummary), all of which are detailed in Appendix~\ref{sec:appendix_a}.
Since segments labeled "None" primarily consist of non-informative responses such as "Yeah" or "Uh-huh," and our objective is to guide the system to generate accurate and meaningful responses using DA tags, we selectively included only those training instances where the gold-standard responses were not labeled "None" in this study. Additionally, we employed text-based human transcriptions rather than audio recordings for our research.

% \begin{table}[t!]
%   \caption{Specific DA tags of operator}
%   \label{tab:dialog_tag}
%   \centering
%   \begin{tabular}{|l|l|}\hline
% DirectionQuestion & SeasonQuestion \\ \hline 
% PeopleQuestion  &  AgeQuestion \\ \hline
% ExperienceQuestion  &  RequestQuestio \\ \hline 
% SearchAdvice  &  RequestConfirm  \\ \hline 
% ChecklistConfirm  &  AddChecklist \\ \hline 
% TravelSummary  &  SearchInform \\ \hline 
% PhotoInform  &  SearchConditionInform \\ \hline 
% NameInform  & IntroductionInform \\ \hline
% OfficeHoursInform  &  PriceInfor \\ \hline 
% FeatureInform  &  AccessInform  \\ \hline 
% PhoneNumberInform  &  ParkInform \\ \hline
% EmptyInform  &  MistakeInform \\ \hline 
% OnScreenSuggest  &  OnScreenQuestion \\ \hline 
% OperatorSpotImpression  &  SearchResultInform  \\ \hline 
% None &  \\ \hline
%   \end{tabular}
% \end{table}

\subsection{Low-Resource Setting}
\begin{table*}[t!]
\caption{Training data quantity for DA prediction across four splits: MO (Minors-Only), ZS (Zero-Shot), LR (Low-Resource), FR (Full-Resource)}
 \label{tab:5_splits}
 \centering
 \small
 \begin{tabular}{ccccccccc} 
 \hline
  Split & Valid & MO-Valid & MO & ZS & LR & FR & Ours & Test \\
 \hline
 1 & 2,027 & 307 & 1,662 & 21,011 & 22,980 & 26,375 & 26,375 & 6,004 \\  
 2 & 2,027 & 199 & 1,117 & 21,011 & 22,327 & 26,434 & 26,434 & 5,945 \\  
 3 & 2,027 & 262 & 1,578 & 21,011 & 22,851 & 26,712 & 26,712 & 5,667 \\  
 4 & 2,027 & 271 & 1,574 & 21,011 & 22,856 & 26,961 & 26,961 & 5,418 \\  
 \hline
 \end{tabular}
\end{table*}

% 我们在5种设定下进行了实验：Minors-Only，Zero-Shot，Low-Resource，Full-Resource，Low-Resource+Augmentation(Ours)。为了模拟特定用户人群的低资源场景，在20名未成年人中，我们仅使用3名未成年人的对话数据用于训练，即18个对话。并使用10名未成年人的60个对话进行评估。
% Minors-Only：仅使用3名未成年人的18个对话。
% Zero-Shot：使用所有的成人与老人的210个对话。
% Low-Resource：使用Minors-Only中3名未成年人的18个对话，以及所有成人与老人的210个对话，共计使用228个对话。
% Full-Resource：使用包括低资源设定中3名未成年人在内的，共10名未成年的60个对话，以及所有成人与老人的210个对话，共计使用270个对话。
% Low-Resource+Augmentation(Ours)：使用Low-Resource中的228个对话，并利用提出的框架增强数据。为了和Full-Resource进行直接的对比，生成数据直到达到与Full-Resource相同的数据量。

We trained five DA prediction models using datasets of varying scales: Minors-Only, Zero-Shot, Low-Resource, Full-Resource, and Low-Resource+Augmentation(Ours). To simulate low-resource conditions for specific user demographics, we used dialogue data from only 3 minors out of a group of 20, totaling 18 dialogues for training. For evaluation, we used 60 dialogues from 10 minors.

\begin{itemize}
\item \textbf{Minors-Only}: Employed only 18 dialogues from 3 minors.

\item \textbf{Zero-Shot}: Utilized all data from adults and seniors, amounting to 210 dialogues.

\item \textbf{Low-Resource}: Combined the 18 dialogues from the Minors-Only with all 210 dialogues from adults and seniors, totaling 228 dialogues.

\item \textbf{Full-Resource}: Included dialogues from 10 minors (60 dialogues), encompassing those from the 3 minors in the low-resource setting, plus all 210 dialogues from adults and seniors, totaling 270 dialogues.

\item \textbf{Low-Resource + Aug(mentation) (Ours)}: Used the 228 dialogues from the Low-Resource and supplemented them using our proposed augmentation framework. Additional data was generated until the dataset size matched that of the Full-Resource for a direct comparison.

\end{itemize}

\subsection{Setup and Details}

% 在提取说话人风格时，我们向GPT-4-0125-preview中input了$m=6$对话，其中3个对话分别来自低资源设定下的3个未成人，另外3个分别来自3名成年人或老人。在生成对话时，我们使用了 GPT-3.5-turbo-0125。

In the process of extracting speaker styles, we fed $m=6$ dialogues into GPT-4-0125-preview, where three were from minors in a low-resource setting, and the other three involved different adults or seniors. For generating training dialogues, GPT-3.5-turbo-0125 was employed.

% 在DA历史生成阶段，我们使用了T5-Large作为PLM。我们进行两次微调，以确保模型在能够生成接近真实人类对话的说话行动历史的基础上，生成符合和具有独特说话策略的未成年人交互时的operator的DA历史。在第一次训练时的学习率为0.0001，第二次仅使用未成年数据训练时学习率为0.00005。生成对话行为历史时，我们将来自成年人和老年人的共210个对话中120个用于训练，90个用于生成。而低资源下的18个未成年人对话均用于训练和生成。为了保证多样性，生成DA历史时设置num_return_sequences=3，即对于一个数据点一次生成3个DA历史。预测当前回合的行动时使用GPT-NeoX，学习率为0.0001，我们为每个模型训练两个epoch。

During the DA history generation phase, we utilized Japanese T5-Large\footnote{https://huggingface.co/retrieva-jp/t5-large-long} as the PLM. We conducted two rounds of finetuning to ensure the model is capable of generating DA histories that not only closely mimic real human conversations but also align with the unique conversational strategies of minors during interactions. During the first training phase, the learning rate was set at 1e-4, and for the subsequent phase exclusively involving data from minors, it was set at 5e-5. 
We utilized 210 adult and elderly conversations for generating DA histories, dividing them into 120 for training and 90 for generation purposes. To ensure data diversity and novelty, we retained only those $(a_t,H_a^{Aug})$ combinations that had not previously existed; all 18 dialogues from 3 minors were included in both training and generation phases. To ensure diversity, we set the num\_return\_sequences=3 when generating DA histories, meaning that for each data point, three DA histories are generated simultaneously.

% 在DA预测阶段，我们分别使用了GPT-NeoX和T5-base作为PLM，以验证生成数据的有效性。我们在同样的DA预测任务下重新构建了训练和评估集，并进行了超参数最适化，具体细节描述在附录B。关于训练，验证集的分配，除Minors-Only之外的所有设定的验证集都相同，来自成年和老年人的21个对话。而Minors-Only的验证集为3个对话，分别来自低资源下的3个未成年人。我们将前三个回合的context和下一回合的DA作为一个数据点。为了验证我们方法的通用性，进行了4个split。每个split使用不同的3名未成年人数据作为低资源设定下的训练数据，同时也更换test数据。去除掉黄金答案为“None”后的各split的数据点统计结果如表1所示。

In the DA prediction phase, Japanese T5-base\footnote{https://huggingface.co/retrieva-jp/t5-base-long} and Japanese GPT-NeoX\footnote{https://huggingface.co/stockmark/gpt-neox-japanese-1.4b} were used as the PLMs to validate the effectiveness of the generated data. We reconstructed the training and evaluation sets for the same DA prediction task to optimize hyperparameters, with specific details provided in Appendix~\ref{sec:appendix_b}.
Regarding the distribution of training and validation sets, the validation sets for all settings, except Minors-Only, are identical, comprising 21 dialogues from adults and seniors. The Minors-Only validation set consists of 3 dialogues from minors in the low-resource scenario. 
To validate the generalizability of our method, we conducted experiments across four splits, each using data from three different minors for training under a low-resource setting, while also varying the test data. 
Details on the data points for each split, after removing entries with a gold-standard answer of "None," are outlined in Table~\ref{tab:5_splits}.

% 考虑到一个语段可能由多个片段组成，而每个片段都可能包含不同的 DA，因此当前回合的黄金DA标签可能不止一个。我们采用了精确匹配率和部分匹配率作为评价指标。完全一致率要求模型预测的标签集与真实的黄金标签集完全一致。这是一个非常严格的指标，能够衡量模型在完全理解对话上下文和精确预测多个相关DA标签方面的能力。而部分一致率评估模型在预测部分正确标签时的表现。这个指标对模型的评估更为宽容，考虑到实际对话中，即使不是每个标签都被精确预测，只要捕捉到了对话的主要意图或行为，也是有实际价值的。因此部分一致率有助于了解模型在实际使用中的鲁棒性。综上所述，这两个指标结合起来可以为评估模型的DA预测能力提供一个平衡严格性和灵活性的方法，从而更准确地反映模型在实际应用中的表现。

Considering that a single utterance may consist of multiple segments (see Figure~\ref{fig:pic_intro} and Figure~\ref{fig:our_model}), each potentially be labeled with a different DA, there may be more than one gold-standard DA label for the current turn. Therefore, we employed both \textbf{exact match and partial match rates} as evaluation metrics. 
The exact match rate is a strict metric requiring the predicted set of labels to completely align with the true set of gold labels, measuring the model’s ability to fully grasp the dialogue context and predict all relevant DA labels accurately. The partial match rate assesses the model's performance in predicting some correct labels. This metric is more lenient, recognizing that in real conversations, capturing the main intent or action of the dialogue, even if not every label is precisely predicted, is still valuable. Therefore, the partial match rate helps understand the model's robustness in practical use. Combined, these two metrics offer a balanced approach to evaluating the model's DA prediction capabilities, providing a more accurate reflection of the model's performance.

\begin{table*}[t!]
\caption{Results across four different splits.}
 \label{tab:result_each}
 \centering
 \small
 \begin{tabular}{|c|l|cc|cc|} \hline
 Split & Setting & \multicolumn{2}{c|}{Japanese GPT-NeoX} & \multicolumn{2}{c|}{Japanese T5-base} \\
  & & Exact Match & Partial Match & Exact Match & Partial Match \\ \hline
 1 & Minors-Only & 0.2451 ± 0.0117 & 0.3447 ± 0.0131 & 0.2533 ± 0.0083 & 0.3519 ± 0.0090 \\  
  & Zero-Shot & 0.2966 ± 0.0071 & 0.4049 ± 0.0092 & 0.3000 ± 0.0059 & 0.4066 ± 0.0053 \\  
  & Low-Resource & 0.3041 ± 0.0070 & 0.4228 ± 0.0073 & 0.3085 ± 0.0065 & 0.4232 ± 0.0064 \\  
  & Low-Resource + Aug (Ours) & \textbf{0.3137 ± 0.0064} & \textbf{0.4320 ± 0.0094} & \textbf{0.3148 ± 0.0050} & \textbf{0.4244 ± 0.0056} \\  \hline
  & Full-Resource & 0.3190 ± 0.0074 & 0.4489 ± 0.0049 & 0.3125 ± 0.0029 & 0.4418 ± 0.0023 \\  \hline \hline 
  2 & Minors-Only & 0.2302 ± 0.0103 & 0.3677 ± 0.0105 & 0.2419 ± 0.0050 & 0.3311 ± 0.0079 \\  
  & Zero-Shot & 0.3162 ± 0.0069 & 0.4247 ± 0.0099 & 0.3200 ± 0.0039 & 0.4263 ± 0.0046 \\  
  & Low-Resource & 0.3220 ± 0.0071 & 0.4401 ± 0.0051 & 0.3257 ± 0.0019 & 0.4430 ± 0.0066 \\  
  & Low-Resource + Aug (Ours) & \textbf{0.3290 ± 0.0083} & \textbf{0.4460 ± 0.0111} & \textbf{0.3270 ± 0.0029} & \textbf{0.4473 ± 0.0095} \\  \hline
  & Full-Resource & 0.3294 ± 0.0068 & 0.4526 ± 0.0074 & 0.3339 ± 0.0052 & 0.4486 ± 0.0075 \\  \hline \hline 
  3 & Minors-Only & 0.2329 ± 0.0033 & 0.3291 ± 0.0069 & 0.2528 ± 0.0038 & 0.3499 ± 0.0010 \\  
  & Zero-Shot & 0.2771 ± 0.0053 & 0.3878 ± 0.0075 & 0.2787 ± 0.0054 & 0.3889 ± 0.0054 \\  
  & Low-Resource & 0.2863 ± 0.0055 & 0.4070 ± 0.0019 & 0.2825 ± 0.0036 & 0.4010 ± 0.0156 \\  
  & Low-Resource + Aug (Ours) & \textbf{0.2906 ± 0.0055} & \textbf{0.4077 ± 0.0067} & \textbf{0.2865 ± 0.0042} & \textbf{0.4097 ± 0.0090} \\  \hline
  & Full-Resource & 0.2889 ± 0.0069 & 0.4282 ± 0.0085 & 0.2986 ± 0.0058 & 0.4270 ± 0.0057 \\  \hline \hline 
  4 & Minors-Only & 0.2325 ± 0.0083 & 0.3336 ± 0.0093 & 0.2429 ± 0.0036 & 0.3480 ± 0.0091 \\  
  & Zero-Shot & 0.2900 ± 0.0066 & 0.4041 ± 0.0066 & 0.2947 ± 0.0047 & 0.4056 ± 0.0059 \\  
  & Low-Resource & 0.2925 ± 0.0067 & 0.4098 ± 0.0088 & 0.2983 ± 0.0031 & \textbf{0.4156 ± 0.0120} \\  
  & Low-Resource + Aug (Ours) & \textbf{0.3005 ± 0.0069} & \textbf{0.4254 ± 0.0087} & \textbf{0.3000 ± 0.0056} & 0.4144 ± 0.0096 \\  \hline
  & Full-Resource & 0.3096 ± 0.0049 & 0.4425 ± 0.0098 & 0.3094 ± 0.0073 & 0.4336 ± 0.0019 \\  \hline 
 \end{tabular}
\end{table*}

\section{Results and Analysis}
% 待翻译
% 表2显示了四种不同拆分下，使用 1 到 5 的种子执行了五次后的平均值和标准差。
% 虽然Minors-Only的训练数据全部来自未成年人，但是由于训练数据过少，其表现不如仅使用成年人和老人对话数据训练的Zero-Shot，因此在其他的设定中，我们也使用了全部的成年人和老人对话数据以增强模型的泛化能力。另外，由于Zero-Shot不使用未成年人的对话，因此在4组不同的split上的Zero-Shot的训练数据相同。Zero-Shot在4组split上的表现差异，进一步表明了即使同样都是未成年人群，模型对其适应能力也存在差异，其中第三个split中的未成年人最有难度。
% 在4组split中，通过我们提出的数据增强框架生成训练数据并用于微调的Low-Resource + Aug (Ours)的表现，在T5和GPT-NeoX上的精确匹配率和部分匹配率都超过了Low-Resource，这表明即使在低资源环境下，我们的方法也能成功捕捉到未成年人说话者的特征，并生成符合未成年人说话行为的对话流程，进而引导训练用对话的生成。然而，尽管在每个split中我们都增强到和Full-Resource相同的数据数量，但大多数情况下Full-Resource的表现最好。可能的原因是我们并没有进行质量控制，因此没有对不好的数据进行筛选，导致同等训练数据量下的适应效果不如所有数据均来自真实人类对话的Full-Resource。另外，由于我们使用的数据集是基于视频通话收集的，即便是人工转录，其中也包含大量填充词和其他形式的口语特征，而ChatGPT更倾向生成流畅的对话，这种差异也可能是Low-Resource + Aug (Ours)不如Full-Resource的原因。

Table~\ref{tab:result_each} shows the mean and standard deviation after five runs using seeds ranging from 1 to 5 across four different splits. 
While the \textbf{Minors-Only} solely comprised data from minors, its performance was inferior to the \textbf{Zero-Shot} model trained only with adult and elderly dialogue data due to the limited amount of training data. Therefore, we also used all available adult and elderly dialogue data in other setups to enhance the model's generalization capabilities. 

Additionally, since \textbf{Zero-Shot} does not use minor's dialogues, the training data remains consistent across the four different splits. The variation in \textbf{Zero-Shot}'s performance across the splits further underscores the differences in the model's adaptability to different minors, with the third split proving most challenging. 

Across the four splits, the performance of our proposed data augmentation framework, \textbf{Low-Resource + Aug (Ours)}, almost all surpassed that of \textbf{Low-Resource} on both T5 and GPT-NeoX in terms of mean exact and partial match rates. This demonstrates that even in a low-resource setting, our method successfully captures the characteristics of minor speakers and generates dialogue flows that align with minor speaking behaviors, thereby guiding the generation of training dialogues.

However, even though we augmented the data to match the quantity of the \textbf{Full-Resource} in each split, \textbf{Full-Resource} typically showed superior performance. A possible explanation is the lack of quality control, which meant that subpar data was not filtered out, leading to poorer adaptation compared to \textbf{Full-Resource}, which used data exclusively from real human conversations. 
Additionally, the "Travel Agency Task Dialogue Corpus," derived from video calls and manually transcribed, may contain colloquial filler words and other informal elements in its complete utterances. In contrast, ChatGPT-generated dialogues tend to be more structured and fluid. This stylistic difference could also contribute to the observed performance disparity between \textbf{Low-Resource + Aug (Ours)} and \textbf{Full-Resource}.

\subsection{Ablation}
\begin{table*}[t!]
\caption{Average results of the ablation experiments across four splits.}
 \label{tab:ablation}
 \centering
 \small
 \begin{tabular}{|l|cc|} \hline
 Setting & Exact Match & Partial Match \\ \hline  
 Low-Resource & 0.3012 & 0.4199 \\  
 w/o DA History Gen & 0.3052 & 0.4263 \\  
 DA History Gen w/o Second Finetune & 0.3072 & 0.4269 \\  
 w/o Speaker Style & 0.3027 & 0.4274 \\  
 Ours & \textbf{0.3085} & \textbf{0.4278} \\ \hline 
 
 \end{tabular}
\end{table*}

% 为了验证提案手法中部分组件的有效性，我们使用Japanese GPT-NeoX在4个split上进行了消融实验。
% w/o DA History Gen : 不生成新的对话行动历史，从Low-Resource中随机抽取现有的对话行动历史用于生成对话数据。
% DA History Gen w/o Second Finetune : 在这一变体中，在训练 DA 历史生成模型时，只进行了一次微调，而不针对未成年人进行第二次微调。
% w/o Speaker style : 在使用LLM生成训练用对话数据时，输入和Ours完全相同的DA历史，但在prompt中不使用抽取到的说话人特征。

% 表4显示了在4个split上的平均结果，我们同样设置种子值为1到5，为每种模型在每个split上进行5次训练。我们发现，w/o DA History Gen和w/o Speaker style的完全一致率，部分一致率都超过了Low-Resource，这表明了特征抽取组件和对话行动生成组件单独使用时生成的训练数据也可以带来性能提升。
% 另外，DA History Gen w/o Second Finetune虽然在训练DA历史生成模型时没有使用目标用户群体的数据进行二次微调，但它的性能超过了w/o DA History Gen。这表明在生成DA历史时，即使没有使用二次微调来优化PLM来适应未成年人，使用全部数据训练的PLM生成的新的$(a_t,H_a)$ combinations也可以带来性能提升。最后，Ours的完全一致率，部分一致率最高，这表明说话人特征抽出和DA历史生成组合起来效果最好，且说明了在训练DA历史生成模型时针对目标年龄段用户二次微调的必要性。

To evaluate the individual effectiveness of components in our proposed framework, we conducted ablation experiments using Japanese GPT-NeoX across four splits:

\begin{itemize}
\item \textbf{w/o DA History Gen}: In this model, we omitted the generation of new DA histories and instead randomly selected DA histories from the Low-Resource for data generation.

\item \textbf{DA History Gen w/o Second Finetune}: This variant involved finetuning the DA history generation model only once, without a second round of finetuning tailored specifically for minors.

\item \textbf{w/o Speaker Style}: This model utilized the same DA histories as our complete method but did not use extracted speaker styles during dialogue data generation.
\end{itemize}

Table~\ref{tab:ablation} shows the average results across the four splits, conducting five trainings for each model in every split with seed values set from 1 to 5. The findings indicate that both \textbf{w/o DA History Gen} and \textbf{w/o Speaker Style} variants achieved higher mean exact and partial match rates than the \textbf{Low-Resource}. This demonstrates that the training data generated through the independent use of style extraction and DA history generation components can also significantly improve performance.

Furthermore, although \textbf{DA History Gen w/o Second Finetune} did not use data from the target user group for a second fine-tuning during the training of the DA history generation model, its performance still surpassed that of \textbf{w/o DA History Gen}. This indicates that in generating DA history, even without a second finetuning to optimize the PLM for minors, the new $(a_t, H_a^{Aug})$ combinations generated by a PLM trained with all available data can still enhance performance. 
Ultimately, \textbf{Ours} achieved the highest rates for both exact and partial matches, indicating that the combination of speaker styles extraction and DA history generation is most effective and underscores the necessity of targeted age-specific second finetuning when training the DA history generation model.

% The results of these ablation experiments are presented in Table~\ref{tab:ablation}. It was observed that each variant surpassed the performance of the Low-Resource setting, indicating that both the speaker feature extraction component and the DA history generation component can independently contribute to performance enhancement.

% The results of these ablation experiments are displayed in Table~\ref{tab:ablation}. We observed that each variant outperformed the Low-Resource setting, indicating that both the speaker feature extraction component and the DA history generation component can independently contribute to performance improvement. 

% We also noted that among these variants, \textbf{w/o Speaker Feature} had the lowest exact match rate, while \textbf{w/o DA History Gen} had the lowest partial match rate. This suggests that speaker features have a greater impact on exact match rate, whereas the generation of new dialogue act history has a more significant effect on partial match rate.

\subsection{Why did the Speaker Style work?}
\begin{figure*}[t!]
  \centering
  \includegraphics[width=0.85\linewidth]{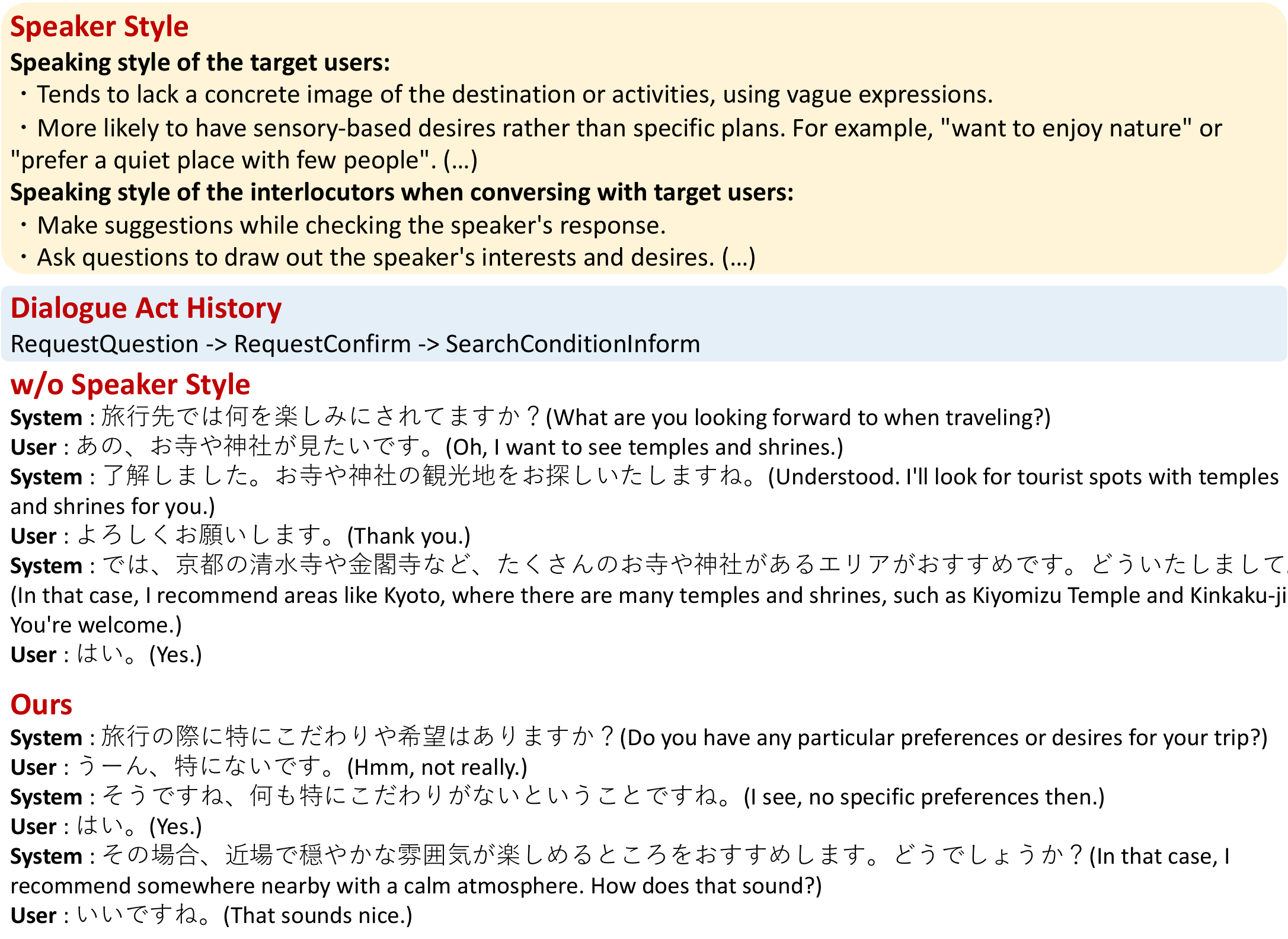}
  \caption{Dialogues generated by the variant without speaker styles and our approach.}
  \label{fig:f_ablation}
\end{figure*}

% 图5显示了在使用相同的DA历史时w/o Speaker Feature和Ours生成的对话。DA历史内容为，先询问用户的关于旅行的request（RequestQuestion），然后确认request（Request Confirm），然后向用户表明自己即将search的内容（SearchConditionInform）。我们可以观察到，由于没有使用Speaker Feature，w/o Speaker Feature的用户给出了具体的旅行要求，对话的展开也很顺利。相比之下，Ours的用户则没有一个清晰的意图，这表明Speaker Feature work，由此产生的对话更符合未成年人的说话特征，也更接近真实的对话场景。

Figure~\ref{fig:f_ablation} displays dialogues generated by \textbf{w/o Speaker Style} and \textbf{Ours}, using the same DA history. The DA history consists of first asking the user a travel-related request (RequestQuestion), then confirming the request (RequestConfirm), and finally indicating the content to be searched (SearchConditionInform). 
We observed that without the speaker style, the user in the \textbf{w/o Speaker Style} provided specific travel requirements, and the dialogue progressed smoothly. In contrast, the user in the \textbf{Ours} did not exhibit a clear intent. This indicates that the speaker style is effective, resulting in dialogues that more closely match the speaking styles of minors and aligning more closely with real human conversations.

\subsection{Why did the DA History Generation work?}

% 我们对比了DA History Gen w/o minor和Ours关于生成DA历史的表现。
% 为了直接比较，我们使用相同的来自90个成年人和老人的对话中的9999个由DA和utterance组成的数据点进行生成，在top_k=50, top_p=0.9, temperature=0.9的设定下为每个数据点生成3个DA历史。去掉重复的DA历史后，DA History Gen w/o minor共生成了7677个新的DA History，Ours生成了10412个。
% 我们计算了生成的DA历史真实存在在17名未成年人（Low-Resource的3人除外）的对话中的数量。结果为DA History Gen w/o minor为908个，而Ours为956个。结合表3我们可以推断，相比使用现有的DA历史的w/o DA History Gen，DA History Gen w/o minor和Ours都生成了符合目标用户的新的DA，这带来了性能的提升。而且Ours针对目标用户进行了二次微调，这使得其生成的DA历史更接近目标用户人群。

We compared the performance in generating DA histories between \textbf{DA History Gen w/o Second Finetune} and \textbf{Ours} on split 1. 

For a direct comparison, we used 9,999 data points $(a_t,S_t)$ from dialogues involving 90 adults and seniors to generate DA histories $H_a^{Aug}$, resulting in three DA histories per data point. This generation was conducted under the settings of top\_k=50, top\_p=0.9, and temperature=0.9.
After removing duplicate $(a_t,H_a^{Aug})$, \textbf{DA History Gen w/o Second Finetune} produced 7,677 new $(a_t,H_a^{Aug})$, whereas \textbf{Ours} generated 10,412.
We assessed how many of these combinations appeared in dialogues involving 17 minors (excluding those from the \textbf{Low-Resource}), finding 908 for \textbf{DA History Gen w/o Second Finetune} and 956 for \textbf{Ours}. Referencing Table~\ref{tab:ablation}, we can infer that compared to \textbf{w/o DA History Gen} which relied solely on existing DA histories, both \textbf{DA History Gen w/o Second Finetune} and \textbf{Ours} generated DAs that were present in the target user group, leading to improved performance. Notably, \textbf{Ours}, which underwent secondary finetuning for the target users, produced more DA histories closely aligned with the target group, enhancing performance.

% % 我们对比了DA History Gen w/o minor和Ours。由于Ours在生成对话行为历史时也使用了未成年对话，因此生成的数量要比DA History Gen w/o minor多。为了直接比较，我们追加了DA History Gen w/ minor，它使用Ours的对话行为历史生成模型，与DA History Gen w/o minor使用同样的输入并输出相同数量的对话行为历史。

% We compared \textbf{DA History Gen w/o minor} with \textbf{Ours}. Since \textbf{Ours} also uses minor dialogues when generating DA history, it produces more instances than \textbf{DA History Gen w/o minor}. 
% To facilitate direct comparison, we added \textbf{DA History Gen w/ minor}, which employs the same DA history generation model as \textbf{Ours}, using the same input as \textbf{DA History Gen w/o minor} and outputting an equivalent number of DA histories.

% % 图5显示了在不同参数设置下，生成的对话行为历史中真实存在在17名未成年人（Low-Resource的3人除外）的对话中的数量。结果表明，在所有设置中，Ours的性能都要好于DA History Gen w/o minor。另外，top_p和temperature的变化对模型的影响要更大。

% Figure~\ref{fig:act_ablation} shows the number of DA histories generated under different parameter settings that actually exist in the dialogues of 17 minors (excluding the 3 from Low-Resource). The results show that in all settings, \textbf{Ours} outperforms \textbf{DA History Gen w/o minor}. Additionally, the variations in top\_p and temperature have a more significant impact on the model's performance.

% \begin{figure}[t!]
%   \centering
%   \includegraphics[width=\linewidth]{pic_ablation_act.pdf}
%   \caption{Quantities of DA histories under different settings.}
%   \label{fig:act_ablation}
% \end{figure}

\section{Conclusion}

% 我们介绍了跨多个年龄层的低资源情况下的数据增强手法。并且通过实验证明了提案手法的稳定性以及单个组件的有效性。本次虽然没有探究提案方法最大能够给模型带来多少提升，我们将把它留在我们以后的工作中。

We introduced a data augmentation method designed to enhance the performance of the DA prediction model for users with limited data and unique conversational styles. Our experiments confirmed the reliability of the proposed method and the effectiveness of its components. While this study did not exhaustively explore the full potential for improvement of the proposed method, we plan to further evaluate this aspect in our future work.

\section*{Acknowledgments}

This work was supported by JSPS KAKENHI Grant Number 19H05692.

\bibliography{custom}

\clearpage
\onecolumn
\appendix

\section{DA tags in Travel Agency Task Dialogue Corpus}
% 在本研究中，我们使用了Inaba等人收集带有任务特定对话行为注释的“Travel Agency Task Dialogue Corpus” \cite{inaba-etal-2024-collection}。该数据集针对旅行代理店对话中的operator和customer分别定义了对话行为标签，其中operator有28个，customer有8个。在本研究中只使用了operator的标签，于表4中展示。

In this study, we utilized the "Travel Agency Task Dialogue Corpus" collected by \citet{inaba-etal-2024-collection}, which includes task specific DA annotations. The dataset defines DA tags for operators and customers in travel agency conversations, with 28 tags for operators and 8 tags for customers. In this study, only the operator's tags were used, as shown in Table~\ref{tab:dialog_tag}.

\begin{table}[h]
  \caption{Task Specific Dialogue Act Tags for Operator Segments.}
  \label{tab:dialog_tag}
  \centering
  \small
  \begin{tabular}{l|p{6cm}|p{6cm}}\hline
     Dialogue Act & Description & Example \\ \hline
     \hline
DirectionQuestion & Question on areas for the desired travel & To which destination are you planning to travel? \\ \hline 
SeasonQuestion & Question on the desired season & When will you go?  \\ \hline 
PeopleQuestion & Question about the number of people traveling and their relationships with the customer & How many people are traveling with you? \\ \hline 
AgeQuestion & Question on the age of customers or their companions & How old are your children? \\ \hline 
ExperienceQuestion & Question about the customer's experience & Have you ever been to Osaka? \\ \hline 
RequestQuestion & Question about the tourist spot request & What would you like to do there? \\ \hline 
SearchAdvice & Questions or suggestions related to the tourist spot information retrieval system & Should I look for a restaurant there? \\ \hline 
RequestConﬁrm & Conﬁrmation ofrequests for tourist spots & You want to go to a Spa, don't you? \\ \hline 
DestinationConﬁrm & Conﬁrmation of destination & Am I correct in Assuming that you are going to Yashi Park? \\ \hline 
AddDestinationList & Addition to destination list by operator & I'll add this location to the list. \\ \hline 
TravelSummary & Summary of trip planning & Looking back, you plan to visit the Toshogu Shrine ﬁrst. \\ \hline 
SearchInform & Operator's declaration of intent to search tourist spots in the system & I will now search. \\ \hline 
PhotoInform & Provide information on photos displayed on the system & Here is a picture of a meal containing a lot of salmon roe. \\ \hline 
SearchConditionInform & Provide information on search conditions & I can also ﬁlter by the time required. \\ \hline 
NameInform & Provide information on the names of tourist spots & There is a commercial complex called the Sapporo Factory. \\ \hline 
IntroductionInform & Provide information on tourist spots based on the system search results & It was established In 1876. \\ \hline 
OﬃceHoursInform & Provide information on hours of operation and closing dates & Our business hours span 10:00 a.m. to 10:00 p.m. \\ \hline 
PriceInform & Provide information on fees and price range & The admission fee is 360 yen. \\ \hline 
FeatureInform & Providing information about the characteristics of tourist spots & It is recommended for women even when it rains. \\ \hline 
AccessInform & Provide information on access & This location is a five-minute walk from the railway station. \\ \hline 
PhoneNumberInform & Provide information on telephone numbers & The phone number is 095 824. \\ \hline 
ParkInform & Provide information on parking & There are three parking lots. \\ \hline 
EmptyInform & Statement that there are no search results or speciﬁc description & I do not see anything in the search results. \\ \hline 
MistakeInform & Correcting errors in tourist spot information & Sorry, this store is open on all days of the week. \\ \hline 
OperatorSpotImpression & Subjective evaluations and assumptions about a tourist spot by operators & This restaurant looks nice and inexpensive. \\ \hline 
SearchResultInform & Report overall search results & It appears there are numerous stores in this location. \\ \hline 
OnScreenSuggest & Suggestions for tourist spots on the shared screen & How about this site? \\ \hline 
OnScreenQuestion & Questions about tourist spots on the shared screen & Which one looks the best, number 1, 2, or 3? \\ \hline 
  \end{tabular}
\end{table}

\label{sec:appendix_a}

% \begin{table}[h]
%   \caption{DA tags of operator.}
%   \label{tab:dialog_tag}
%   \centering
%   \begin{tabular}{|l|l|}\hline
% DirectionQuestion & SeasonQuestion \\ \hline 
% PeopleQuestion  &  ExperienceQuestion \\ \hline
% AgeQuestion  &  RequestQuestio \\ \hline 
% SearchAdvice  &  RequestConfirm  \\ \hline 
% ChecklistConfirm  &  AddChecklist \\ \hline 
% TravelSummary  &  SearchInform \\ \hline 
% PhotoInform  &  SearchConditionInform \\ \hline 
% NameInform  & IntroductionInform \\ \hline
% OfficeHoursInform  &  PhoneNumberInform \\ \hline 
% FeatureInform  &  AccessInform  \\ \hline 
% PriceInform  &  OperatorSpotImpression \\ \hline
% EmptyInform  &  MistakeInform \\ \hline 
% OnScreenSuggest  &  OnScreenQuestion \\ \hline 
% ParkInform  &  SearchResultInform  \\ \hline 
% None &  \\ \hline
%   \end{tabular}
% \end{table}

\section{Hyperparameter Optimization}
\label{sec:appendix_b}

% 我们在实验中进行了超参数优化。
% 为了找到T5-base的最优超参数，我们进行了网格搜索，批量大小为{8, 16, 32, 64}，warmup ratio为{0, 0.1, 0.2}，学习率为{3e-3, 2e-3, 1e-3, 9e-4, 8e-4}。最终的设定为，批量大小设为64，warmup ratio设为0.1，学习率设为1e-3。
% 同样，我们对GPT-NeoX也进行了网格搜索，批量大小为{4, 8, 16}，warmup ratio为{0.1, 0.2, 0.3}，学习率为{3e-4, 2e-4, 1e-4, 9e-5, 8e-5, 7e-5, 6e-5, 5e-5, 4e-5}。最终的设定为，批量大小设为8，warmup ratio设为0.1，学习率设为9e-5。

During our experiments, we performed hyperparameter optimization.

For T5-base, we conducted a grid search with batch sizes of \{8, 16, 32, 64\}, warmup ratios of \{0, 0.1, 0.2\}, and learning rates of \{3e-3, 2e-3, 1e-3, 9e-4, 8e-4\}. The optimal configuration was identified as a batch size of 64, a warmup ratio of 0.1, and a learning rate of 1e-3.

Similarly, for GPT-NeoX, we conducted a grid search with batch sizes of \{4, 8, 16\}, warmup ratios of \{0.1, 0.2, 0.3\}, and a range of learning rates of \{3e-4, 2e-4, 1e-4, 9e-5, 8e-5, 7e-5, 6e-5, 5e-5, 4e-5\}. The best settings were determined to be a batch size of 8, a warmup ratio of 0.1, and a learning rate of 9e-5.

\newpage
\section{Details for Speaker Styles Extraction.}
\label{sec:appendix_c}

% 我们使用图5所示的prompt提示GPT-4-0125-preview提取说话人风格，其中的6个对话分别来自不同的用户，3人来自目标用户群体，另外3人来自非目标用户群体。由于提取时保持默认的温度（即temperature=1），生成的结果具有多样性。我们进行了多次提取，并手动组合了提取到的说话人风格。最终的说话人风格由图6所示，全部用于后续的训练用对话数据生成。

\begin{figure}[h!]
  \centering
  \includegraphics[width=0.65\linewidth]{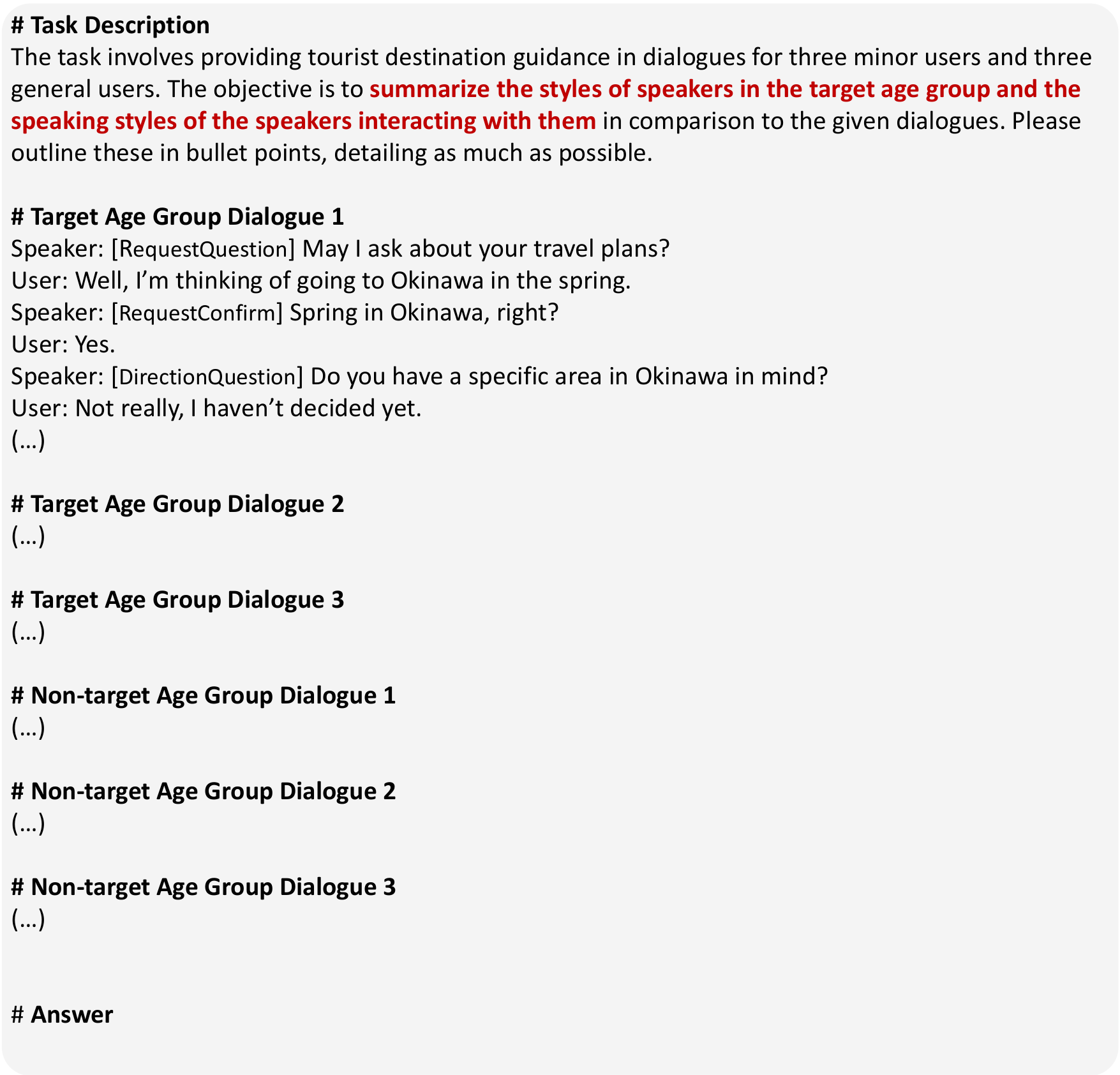}
  \caption{Prompt for Speaker Styles Extraction.}
  \label{fig:pic_prompt_1}
\end{figure}

\begin{figure}[h!]
  \centering
  \includegraphics[width=0.92\linewidth]{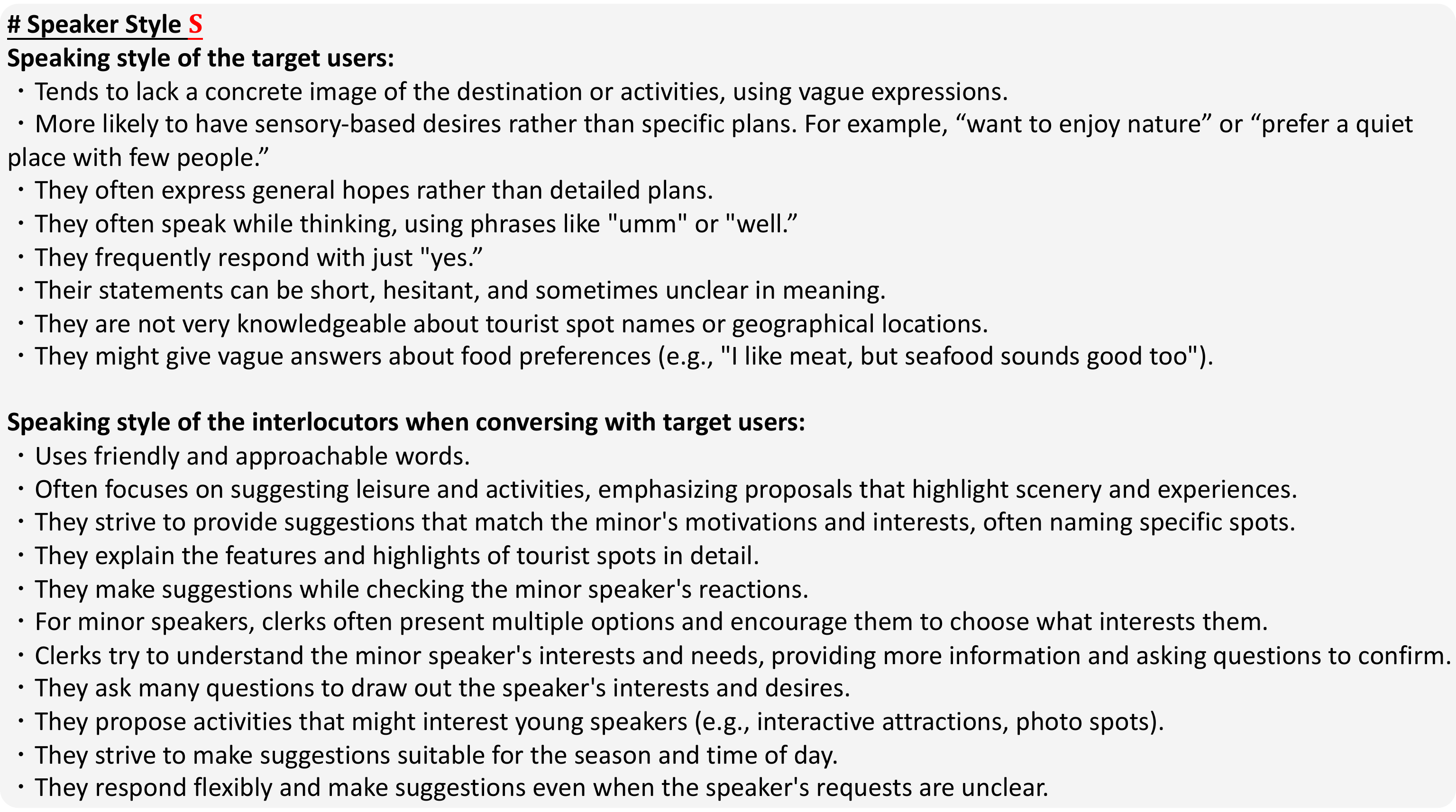}
  \caption{Extracted Speaker Styles. They are utilized for subsequent dialogue generation.}
  \label{fig:pic_prompt_ex}
\end{figure}

We utilized the prompt shown in Figure~\ref{fig:pic_prompt_1} to extract speaker styles using the GPT-4-0125-preview model, with six dialogues from different users, three from the target user group and three from a non-target user group. As the extraction was conducted with the default temperature setting (i.e., temperature=1), the generated results were diverse. We performed multiple extractions and manually combined the extracted speaker styles. The consolidated speaker styles, as illustrated in Figure~\ref{fig:pic_prompt_ex}, were all used for subsequent dialogue data generation.

\newpage
\section{Prompt used for Training Dialogue Generation.}
\label{sec:appendix_d}

% 图7所示的prompt被用于提示GPT-3.5-turbo-0125生成训练用对话数据。我们在prompt中加入了7个exapmles，以控制其生成质量。所有的examples均来自“Travel Agency Task Dialogue Corpus” \cite{inaba-etal-2024-collection}中目标用户群体的真实对话。

\begin{figure}[h!]
  \centering
  \includegraphics[width=1\linewidth]{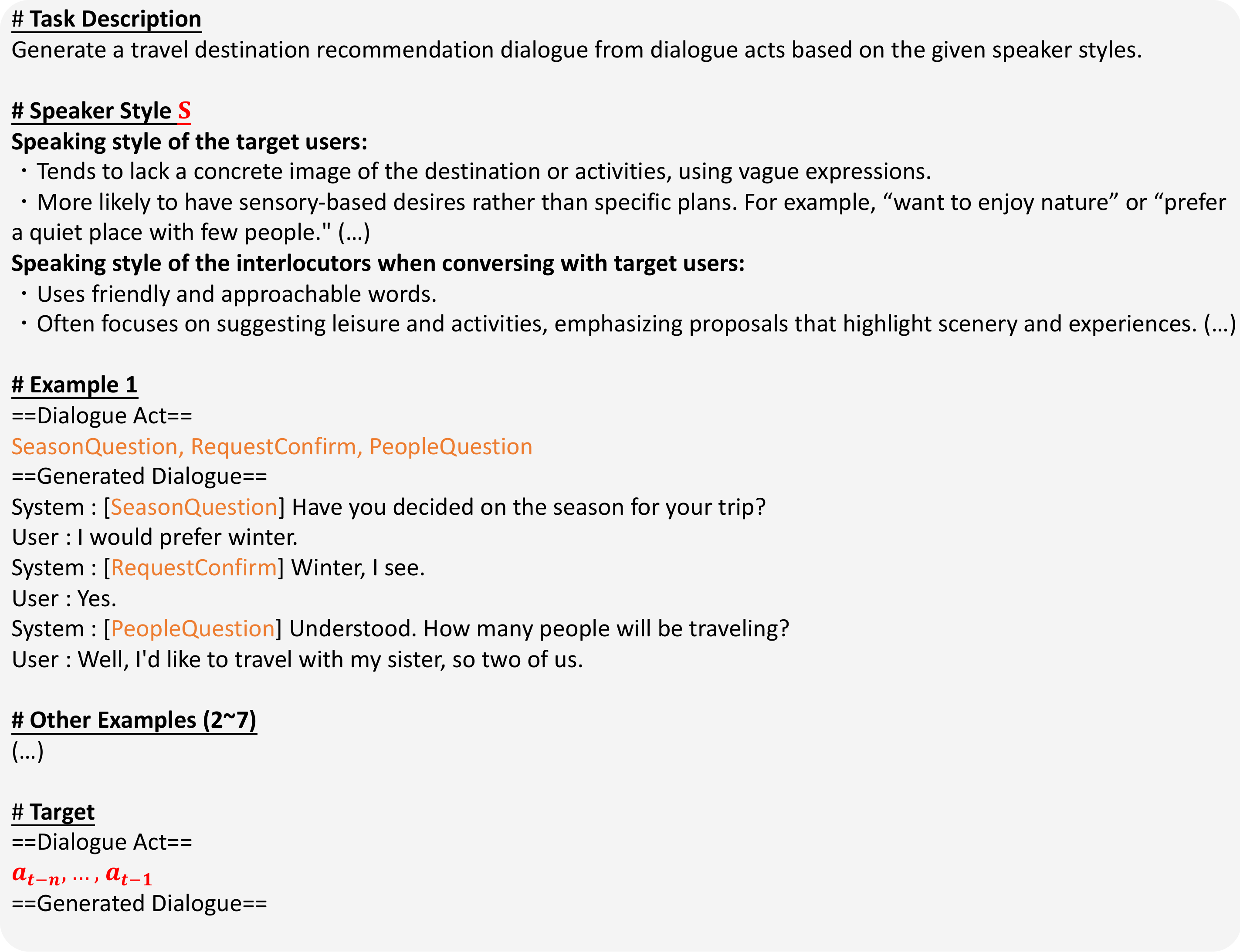}
  \caption{Prompt for Dialogue Generation. Red indicates the condition generated in previous steps.}
  \label{fig:pic_prompt}
\end{figure}

The prompt shown in Figure~\ref{fig:pic_prompt} was employed to instruct GPT-3.5-turbo-0125 to generate dialogue data for training. We included seven examples in the prompt to control the quality of generation. All examples originated from real conversations of the target user group in the "Travel Agency Task Dialogue Corpus" \cite{inaba-etal-2024-collection}.

\end{document}